\documentclass[conference]{IEEEtran}
\IEEEoverridecommandlockouts
\usepackage{cite}
\usepackage{amsmath,amssymb,amsfonts}
\usepackage{algorithmic}
\usepackage{graphicx}
\usepackage{textcomp}
\usepackage{xcolor}
\usepackage[]{footmisc}
\usepackage{hyperref}

\ifCLASSOPTIONcompsoc
    \usepackage[caption=false, font=normalsize, labelfont=sf, textfont=sf]{subfig}
\else
\usepackage[caption=false, font=footnotesize]{subfig}
\fi
\def\BibTeX{{\rm B\kern-.05em{\sc i\kern-.025em b}\kern-.08em
    T\kern-.1667em\lower.7ex\hbox{E}\kern-.125emX}}
\begin{document}

\title{Shapes2Toon: Generating Cartoon Characters from Simple Geometric Shapes}
\makeatletter
\newcommand{\linebreakand}{%
  \end{@IEEEauthorhalign}
  \hfill\mbox{}\par
  \mbox{}\hfill\begin{@IEEEauthorhalign}
}
\makeatother

\author
{\IEEEauthorblockN{Simanta Deb Turja\IEEEauthorrefmark{1}}
\IEEEauthorblockA{
\textit{Ahsanullah University of Science}\\
\textit{and Technology, Bangladesh} \\
simantaturja@gmail.com}
\and
\IEEEauthorblockN{Mohammad Imrul Jubair\IEEEauthorrefmark{1}}
\IEEEauthorblockA{
\textit{University of Colorado Boulder,}\\
\textit{United States}\\
mohammad.jubair@colorado.edu}
\and
\IEEEauthorblockN{Md. Shafiur Rahman}
\IEEEauthorblockA{
\textit{Ahsanullah University of Science}\\
\textit{and Technology, Bangladesh}\\
dipu.shafiur@gmail.com}
\linebreakand 
\IEEEauthorblockN{Md. Hasib Al Zadid}
\IEEEauthorblockA{
\textit{Ahsanullah University of Science}\\
\textit{and Technology, Bangladesh} \\
hzsakkhor@gmail.com}
\and
\IEEEauthorblockN{Mohtasim Hossain Shovon}
\IEEEauthorblockA{
\textit{Ahsanullah University of Science}\\
\textit{and Technology, Bangladesh} \\
mohtasimshovon6352@gmail.com}
\thanks{\IEEEauthorrefmark{1} these authors contributed equally.}
\and
\IEEEauthorblockN{Md. Faraz Kabir Khan}
\IEEEauthorblockA{
\textit{Ahsanullah University of Science}\\
\textit{and Technology, Bangladesh}\\
farazkabir@gmail.com}
}

\maketitle

\begin{abstract}
Cartoons are an important part of our entertainment culture. Though drawing a cartoon is not for everyone, creating it using an arrangement of basic geometric primitives that approximates that character is a fairly frequent technique in art. The key motivation behind this technique is that human bodies---as well as cartoon figures---can be split down into various basic geometric primitives. Numerous tutorials are available that demonstrate how to draw figures using an appropriate arrangement of fundamental shapes, thus assisting us in creating cartoon characters. This technique is very beneficial for children in terms of teaching them how to draw cartoons. In this paper, we develop a tool---\textit{shape2toon}---that aims to automate this approach by utilizing a generative adversarial network which combines geometric primitives (i.e. circles) and generate a cartoon figure (i.e. Mickey Mouse) depending on the given approximation. For this purpose, we created a dataset of geometrically represented cartoon characters. We apply an image-to-image translation technique on our dataset and report the results in this paper. The experimental results show that our system can generate cartoon characters from input layout of geometric shapes. In addition, we demonstrate a web-based tool as a practical implication of our work.
\end{abstract}

\begin{IEEEkeywords}
geometric shapes, circles, image-to-image translation, pix2pix, gan, cartoon.
\end{IEEEkeywords}

\section{Introduction}
\label{sec:introduction}
In our everyday lives, cartoons are a common creative genre. Apart from entertainment pursuits, their uses include anything from publishing in print media to children's educational narrative~\cite{Chen_2018_CVPR}. It is not simple for everyone to create cartoon characters from sketches; rather, we must use a variety of techniques. The most often used method for a beginner is to begin with some basic shapes---circles, ovals, triangles, and rectangles---and then build the character upon them. When we combine these shapes, we get basic designs and forms of cartoon such as heads, bodies, and buildings~\cite{fairrington2009drawing}~\cite{curto2015art}. An overview of the approach is illustrated in Fig.~\ref{fig:draw_overview}. The primary reason behind this method of using basic geometric shapes is that---things we sketch are fundamentally composed of basic forms. Looking around, we can see that everything, including humans, can be broken up into different geometric shapes. By using these primitives, we can not only produce stylistic proportions but also give our characters apparent personalities ~\cite{cole_2021}. A competent cartoonist is naturally adept at extracting the essential elements of an object or person and representing them with basic shapes~\cite{cabral_2021}. Moreover, this method is often used in cartoon sketching lessons. The novices---as all artists do---begin with studying the basic forms and expressions, then go to the little nuances and quirks that add to the enjoyment of an expression~\cite{making_faces}. Therefore, fundamental geometric forms arguably play a major part in the creation of cartoon characters.

While the vast majority of individuals are incapable of drawing a cartoon figure perfectly or in the manner of a professional artist, anyone may make a structure of the cartoon character using geometric shapes such as circles and ellipses (as shown in Fig.~\ref{fig:draw_overview}). The approach of building the cartoon character based on shapes is very effective for children, since the understanding of geometric shapes is being developed in childhood, essentially based on the assessment of their responses in the course of tasks associated with recognition, classification, or explanation of geometric models~\cite{clements2004geometric}. Children can design drawings of characters, as well as numerous articulations of characters, by just sketching a few shapes~\cite{article}.
\begin{figure}[ht]
  \centering
  \includegraphics[width=\linewidth]{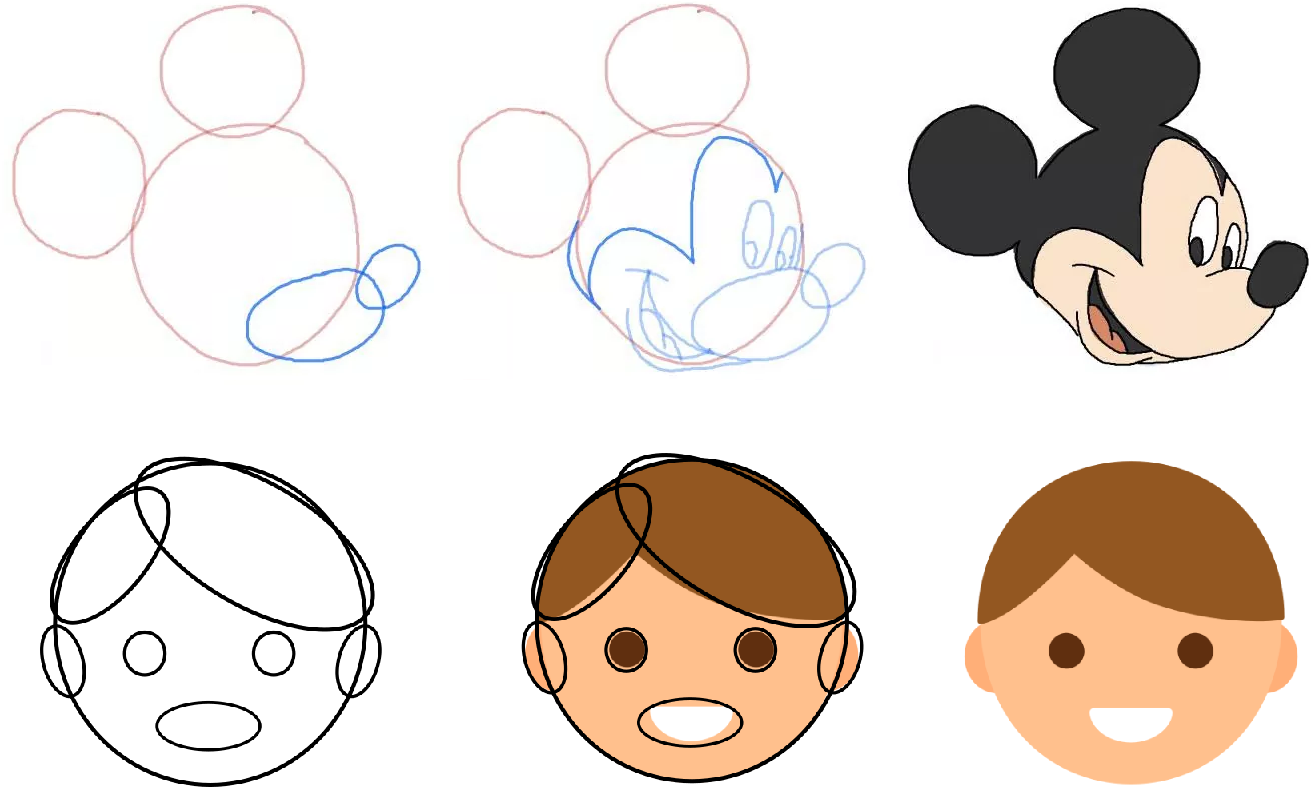}
  \caption{An overview of drawing cartoon using simple geometric shapes. Here (\textit{top row}), we first sketch an arrangement of $3$ circles and $2$ ovals (\textit{left}). Which can later be useful to build the cartoon faces (\textit{middle \& right}). The figure is reproduced from~\cite{how_to_draw_2021}. Another example for a different cartoon is shown in \textit{bottom row}.}
  \label{fig:draw_overview}
 \end{figure}

Our paper focuses on this sketching method and attempts to automate it using a Computer Vision technology. The recent breakthrough of the Generative Adversarial Network~\cite{Goodfellow-et-al-2016} has influenced the researchers to work with image-to-image translations and to develop different stunning variations, for instance converting a horse into a zebra~\cite{chu2017cyclegan}, a sketch into image~\cite{chen2018sketchygan}, a stroke to motif~\cite{jamdani2020}, etc.

\subsection*{Contribution}
Being motivated from the above mentioned works, we aim to create a cartoon figure utilizing an image-to-image translation method in our work that accept adjoined geometric shapes from the user and synthesize a cartoon character depending on the arrangement of those shapes. We call this tool \textit{shapes2toon}. We think that our approach will help in revealing the beginner cartoonist's visualizing ability and in teaching children to draw. Moreover, it can benefit a drawing enthusiast and can be used as a sketching tool by anybody, even those with little drawing ability.
Our work has made the following contributions.
\begin{itemize}
  \item{We propose an approach--\textit{shapes2toon}--that converts an arrangement of geometric shapes to a cartoon character. As an application of our approach, we regard recreating the renowned Disney\footnote{https://thewaltdisneycompany.com} cartoon character Mickey Mouse based on the user's geometric input of shapes. We limit ourselves to circles and ovals when it comes to basic shapes. For this purpose, we created a dataset that includes cartoon characters and their representations in geometric forms. We developed a Javascript-based tool to manually trace the geometrical approximation of the cartoon characters to build the dataset.
  Moreover, we make this dataset available to the community for future research and can be found here: {\textit{https://tinyurl.com/shapes2toon-dataset}}.
  }
  \item{We presented the results of applying a state-of-the-art image-to-image translation technique---\texttt{pix2pix}~\cite{isola2017image}---on our dataset. Additionally, we created a tool that allows users to draw geometric objects on a digital canvas and our system would generate the appropriate Mickey Mouse(see fig.~\ref{fig:web_tools}}).
\end{itemize}

The following describes the structure of this paper. Section~\ref{sec:related} discusses related studies. In Section~\ref{sec:our_dataset}, we describe our dataset followed by an overview of applying image-to-image translation on it in Section~\ref{sec:exp} along the outcomes of our experiment. Section~\ref{sec:con} concludes our discussion by outlining the limits of our research as well as possible future directions.

\section{Background and Related Works}
\label{sec:related}
Our work falls under the field of image-to-image (I2I) translation. Hence, we provide in this section our study on related I2I methods that are pertinent to our research objective and works as our motivation.

\subsection{Generative Adversarial Network (GAN)}
One of the main backbones of I2I is Generative Adversarial Network, or GAN, which was first proposed by \cite{Goodfellow-et-al-2016} can be considered an unconditional GAN. It constructs a structured probabilistic model taking latent noise variables $z$ and observed real data $x$ as inputs. GAN is an adversarial method that is comprised of two neural network models: the generator and the discriminator, each represented as a differentiable function with parameters. Generator $G$ attempts to create convincing false pictures, while discriminator $D$ is taught to differentiate between the two. This game's solution is a Nash equilibrium between the two participants. The objective optimization problem is as shown in Eq.~\ref{eq:gan}~\cite{pang2021imagetoimage}.
\begin{equation}
\begin{aligned}
\mathop{min}\limits_{G}\mathop{max}\limits_{D}{L}(D,G)=E_{x\sim{p_{data}(x)}}[logD(x)]+\\
E_{z\sim{p_{z}(z)}}[log(1-D(G(z)))].
\end{aligned}
\label{eq:gan}
\end{equation}
where $x$ and $z$ denote the real data and random noise vector respectively. $G(z)$ are the fake samples produced by the generator $G$, and $D(x)$ indicates the probability that $D$'s input is real, and $D(G(z))$ is the probability that $D$ discriminates between the input from $G$.

A frequently used variant of GAN is conditional GAN (cGAN)~\cite{mirza2014conditional}, which allows for more control over the output. As a result, the authors suggested concatenating extra information $y$ with $z$ to create picture $G(z|y)$. The conditional input $y$ may be any kind of data, including data labels, text, and picture characteristics.

\subsection{Image-to-Image Translation (I2I)}
I2I can be achieved using concept of GAN. In principle, I2I converts an input picture $x_{P}$ from a source domain $P$ to a target domain $Q$ while preserving the intrinsic source content and transferring the extrinsic target style. We need to train a mapping $G_{P\to{Q}}$ that generates image $x_{PQ}\in{P}$ similar to target image $x_{Q}\in{Q}$ given the input source image $x_{P}\in{P}$. Mathematically, we can model this translation process as shown in Eq.~\ref{eq:pro_setting}~\cite{pang2021imagetoimage}.
\begin{equation}
\label{eq:pro_setting}
x_{PQ}\in{P}: x_{PQ} = G_{P\to{Q}}(x_P).
\end{equation}

Authors in \cite{isola2017image} demonstrate the application of conditional GAN to the I2I domain by proposing \textit{pix2pix} for solving a variety of supervised I2I tasks. Along with the pixelwise regression loss $L_1$ between the translated picture and the ground truth, this method employs adversarial training loss $L_{cGAN}$ controlled by hyperparameter $\lambda$ to guarantee that the outputs are indistinguishable from "actual" images. The objective is shown in Eq.~\ref{eq:pix2pix}.
\begin{equation}
{L} = \mathop{min}\limits_{G}\mathop{max}\limits_{D}{L}_{cGAN}(G,D) +\lambda{L}_{{L}_{1}}(G).
\label{eq:pix2pix}
\end{equation}
In case of pix2pix, paired training samples are used where the input and the ground-truth image domains are aligned. However, \textit{CycleGAN}~\cite{chu2017cyclegan} is a very popular technique for automatically training image-to-image translation models without the use of paired samples. Here, unsupervised learning is used to train the models using a collection of pictures from the source and target domains that are not required to be linked in any way.
In the practical field of computer vision, pix2pix offers a powerful foundation for image translation, inspiring several enhanced I2I efforts.\\

\subsubsection{User's Input Outline based I2I}
Numerous improvements and applications have been created based on the conditional GAN framework. Given that our study is concerned with user input drawings or outlines, we briefly address several relevant studies.\\

To start with, \textit{SPADE}~\cite{park2019gaugan}---also known as \textit{GauGAN}---recommends the use of a spatially adaptive normalizing layer, for photo-realistic and greater quality improvement of the synthesized pictures. SPADE utilizes only one style code to govern the whole style of an image and inserts style information only at the beginning of a network, rather than throughout. The model allows users to control the style and content of synthesis results along with generating multi-modal results.

In \textit{SketchGAN}~\cite{liu2019sketchgan}, the authors proposed a GAN for the completion of sketches. Their approach can be used to complete input drawings from a variety of categories of objects. Their main idea is to jointly conduct sketch completion and recognition tasks.

Another GAN-based end-to-end trainable sketch to image synthesis technique called \textit{SketchyGAN} is proposed in~\cite{DBLP:journals/corr/abs-1801-02753} which can create objects from $50$ different classes and is GAN-based. This algorithm takes as input an item drawing and produces as output a realistic picture of the object in a comparable position to the sketch.

Besides, the authors propose in~\cite{zeng2021strokegan} \textit{StrokeGAN} for generating Chinese fonts from unpaired data. A one-bit stroke encoding is used to capture the mode information of Chinese characters, which is subsequently used to train CycleGAN~\cite{chu2017cyclegan}, reducing mode collapse and increasing character variety. They train CycleGAN with a stroke-encoding reconstruction loss to maintain stroke encoding.\\

\begin{figure*}[ht]
    \centering
    \includegraphics[width=\linewidth]{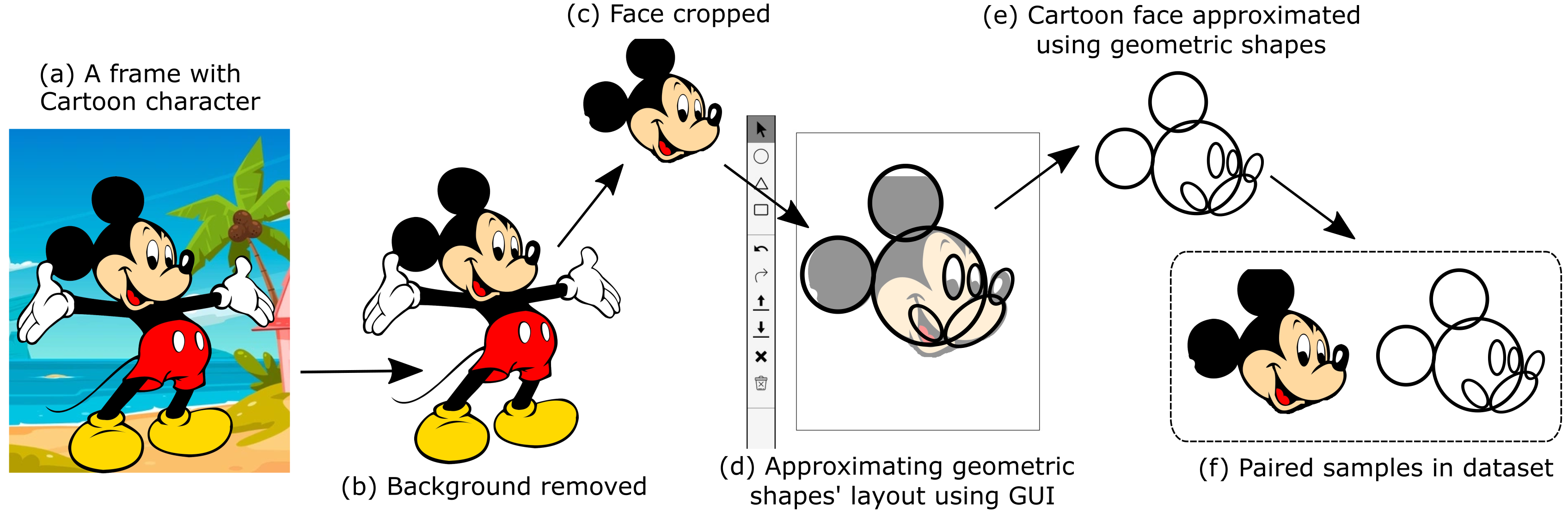}
    \caption{The workflow of building our dataset.}
    \label{fig:create_dataset}
\end{figure*}

\subsubsection{I2I on Cartoon}
Since our paper is concerned with the synthesis of cartoon characters, we provide few existing I2I-based works that focused on the domain of cartoon.\\

Because many renowned cartoon pictures were developed based on real-world settings, the authors in~\cite{Chen_2018_CVPR} were driven to propose \textit{CartoonGAN}. This is an unique GAN-based method to picture cartoonization for transforming photos of real-world scenes into cartoon style images. An unpaired collection of photographs and a set of cartoon pictures are required for training with this approach. 
The reverse task of CartoonGAN is taken into consideration in \textit{toon2real}~\cite{toon2real}, which uses GAN to convert cartoon pictures into photo-realistic ones. They present a technique for picture translation from the cartoon domain to the photo-realistic domain that is based on the CycleGAN model.

The work \textit{Auto-painter}~\cite{liu2017auto} proposed an auto-painter learning model for automatically generating painted cartoon pictures from a sketch using cGANs. Cartoon pictures have more creative color palettes, which may necessitate additional modeling limitations. The generator is trained using constraints such as data variability loss, pixel loss, and feature loss in order to produce more beautiful color collocations. Additionally, the authors provided a color control for the auto-painter, allowing users to paint in their own hues.\\

A number of additional works, such as~\cite{shu2021gan}, \cite{Wang_2020_CVPR}, \cite{back2021finetuning} and~\cite{thakur2021white}, that focus on photo cartoonization have also been produced. The \textit{GanToon}~\cite{GANtoon} generates cartoons which are new in design without any input from the user. The authors considers the popular cartoon character \textit{Tom} as their target. There are additional works that generate Avatar and caricature--- for instances~\cite{roboco}, \cite{jhawar_2021}, \cite{Gong_2020_WACV}, \cite{cao2018cari} etc.---but are vaguely relevant to our research topic, and thus we do not go into them in depth in our literature review.\\

None of the techniques described above is concerned with the generation of cartoons from a layout of fundamental geometric shapes. In this paper we take attempt to accomplish this task using pix2pix as our framework. Our primary challenge in completing this work was the absence of available datasets. As a result, we create a paired dataset for our pix2pix model, which is discussed in further detail in the next section.

\section{\uppercase{Our Dataset}}
\label{sec:our_dataset}
For our~\textit{shapes2toon}, we fix our target character to the famous Mickey Mouse from Walt Disney. In order to reduce complexity, we only focus on the face of the cartoon in this paper. Also, to introduce simplicity to the users, we allow them to input circles and ovals only to approximate the shape of a Mickey Mouse. Hence, we focus on building a dataset in a pair of two categories: (1) the combination of circles and ovals that best approximate Mickey Mouse's details, and (2) the ground truth of the Micky Mouse character. Based on this approach, we build our dataset with total of $7500$ augmented Mickey Mouse images. The steps required for creating the dataset(see fig.~\ref{fig:create_dataset}) are listed below.

\subsection{Image Collection}
Our main source of data collection was YouTube, e.g. \texttt{Micky Mouse YouTube channel}\footnote{https://www.youtube.com/c/MickeyMouse}. We first manually selected the videos that have significance presence of Mickey Mouse, and then extracted frames from those media files. Since you only want to focus on the Mickey Mouse, we cropped out the region of the character's face using image processing softwares.

\subsection{Pre-processing}
\label{subsec:pre}
Background of each samples needs to be removed for better training of our model. However, this step is a tedious process as it involves lots of time and attention to detail---even for experienced designer---via image editing software. Therefore we exploited AI based background removal tool \texttt{remove.bg}\footnote{https://www.remove.bg/} to serve our purpose. We also processed the background removed image in order to keep the facial portion only.

\subsection{Approximating Character with Basic Geometric Shapes}
As we determined to develop a paired dataset, we required the collected samples to be processed further to produce the duad. In this step our purpose was to develop a system that takes the Mickey Mouse character from previous stage (Section~\ref{subsec:pre}) and generates a layout of circles or ovals that resemble the character.
We initially attempted to construct the approximated layout of circles and ovals via Hough transformation for ellipse detection~\cite{houghEllipse}\cite{Xie2002ANE}.
However, due to the complex body forms of cartoon characters, hough transformation
failed in the majority of situations.
In addition, we also applied template matching technique~\cite{hashemi2016template} with adaptive templates constructed from interested regions of Mickey Mouse character. However, the results were quite unsatisfactory in terms of determining where the ROIs were located even for empirical values of thresholds.\\
\begin{figure}[htb]
    \centering
    \includegraphics[width=\linewidth]{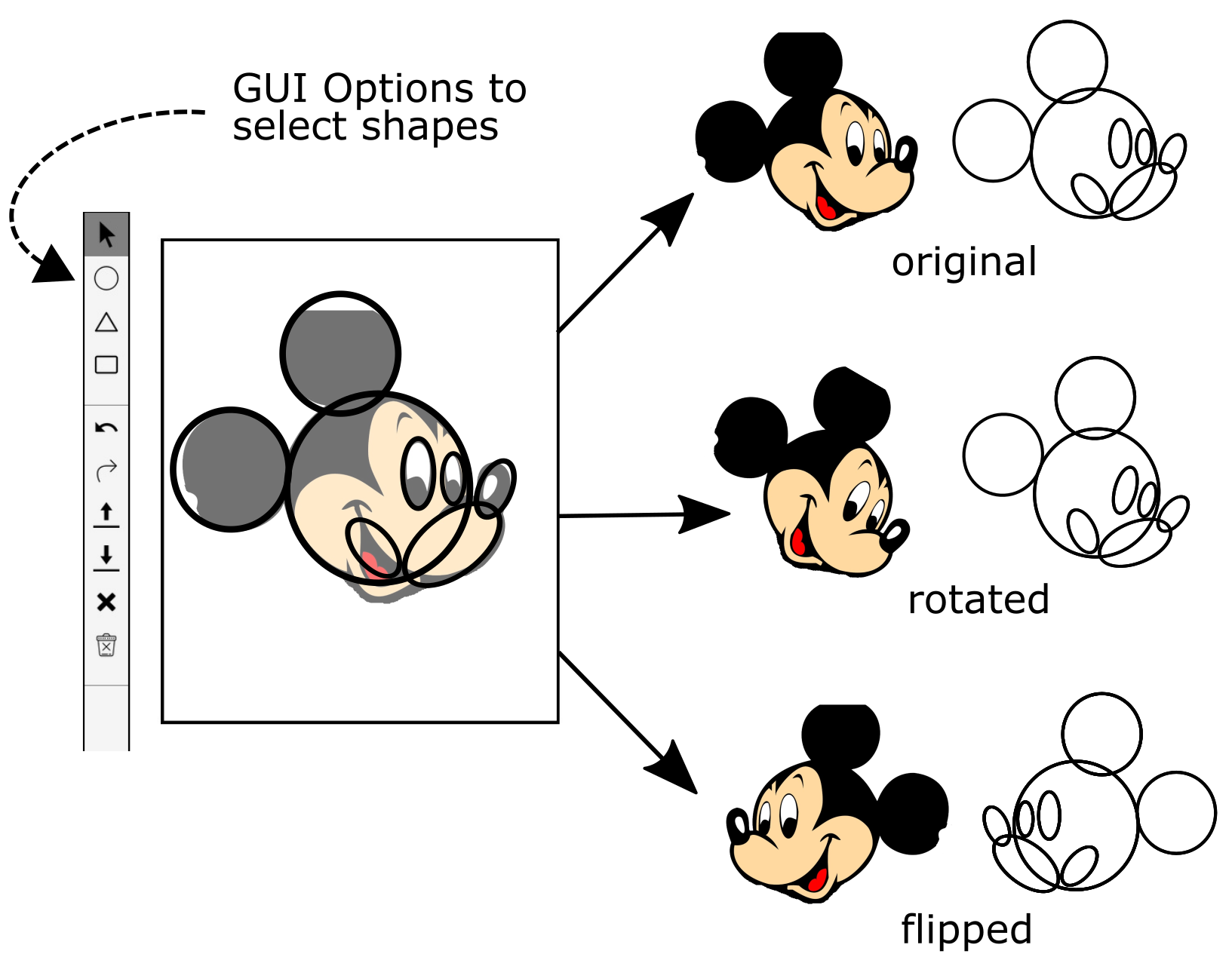}
    \caption{An overview of our tool (\textit{left}) for approximating a cartoon face with basic geometric shapes. Examples of augmented paired outputs are shown in the \textit{right}.}
    \label{fig:toon_shape_tool}
\end{figure}

\subsubsection{Manual Approximation}
Earlier mentioned attempts and observations for extracting geometric feature finally drove us to generate approximated layout of circles and ovals manually. For this purpose we developed a JavaScript based tool (see Fig.~\ref{fig:toon_shape_tool}) and assigned volunteers to use then for completing paired dataset. Using this tool, the volunteers can upload images of Mickey Mouse, navigate through its different options from its GUI and selects desired geometric primitives (i.e. circles) to draw it on top of the character. The layouts of the geometric shapes were stored as a seperate image files. The shapes can be adjusted---such as rotated, scaled, shifted, etc---based on the contents of the uploaded cartoon.
\subsubsection{Data Augmentation}
Using our tool, We created $500$ cartoon images with their geometrical representation. An additional feature of this tool is the integrated image data augmentation~\cite{Shorten2019}. We apply different versions for each paired samples---such as rotated, scaled, flipped, translated---using the feature of our tool to obtain results from the model (see Fig.~\ref{fig:toon_shape_tool}). The final augmented dataset contains $7500$ images of cartoon characters.

\begin{figure}[htb]
    \centering
    \includegraphics[width=\linewidth]{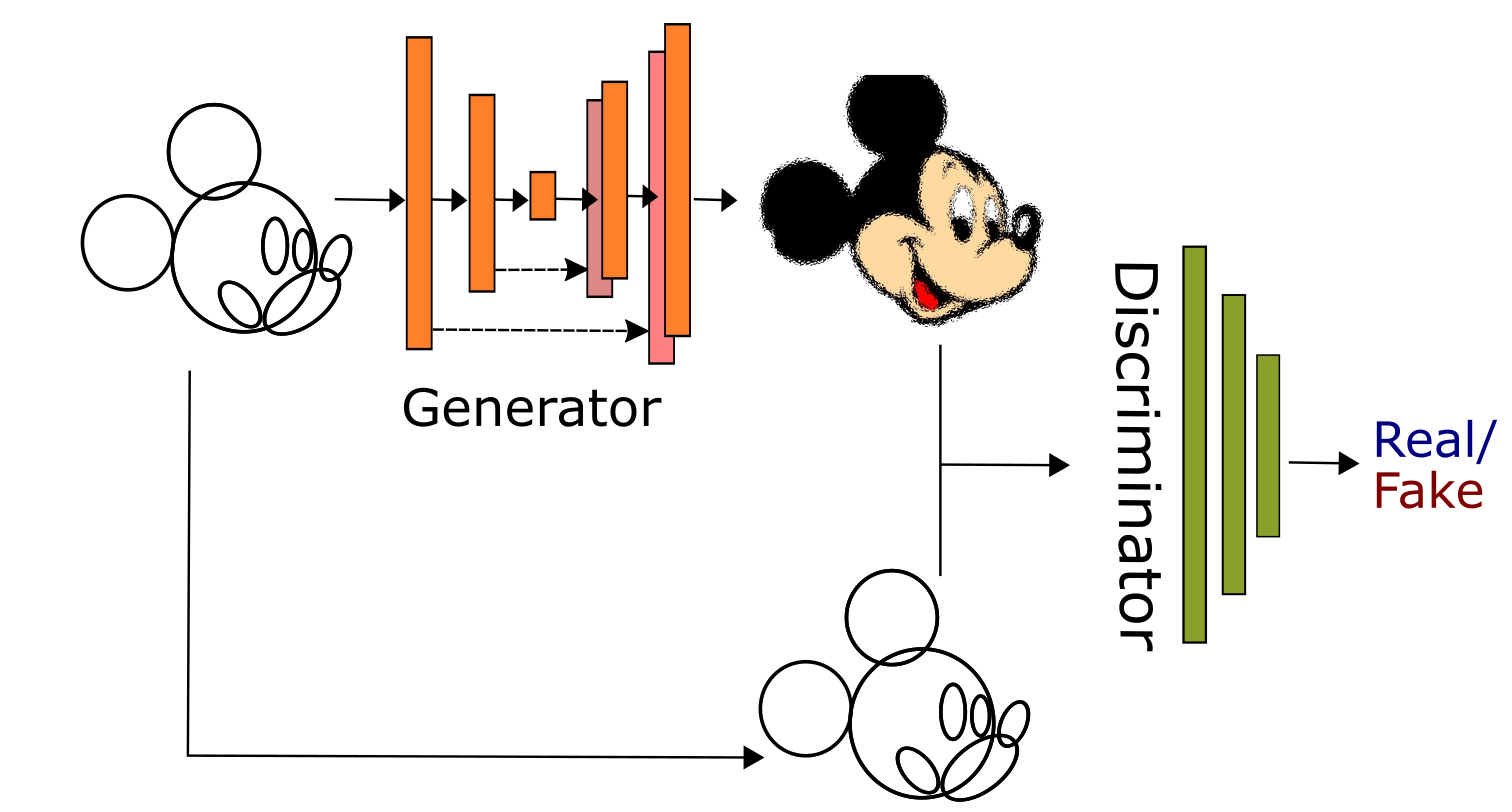}
    \caption{Training of \textit{Pix2pix} to associate shapes with cartoons. The discriminator learns to distinguish between fake (generator-generated) and real \{\textit{shapes, cartoon}\} tuples. The generator learns to fool the discriminator.}
    \label{fig:pix2pix_arch}
\end{figure}

\begin{figure}[htb]
    \centering
    \includegraphics[width=0.95\linewidth]{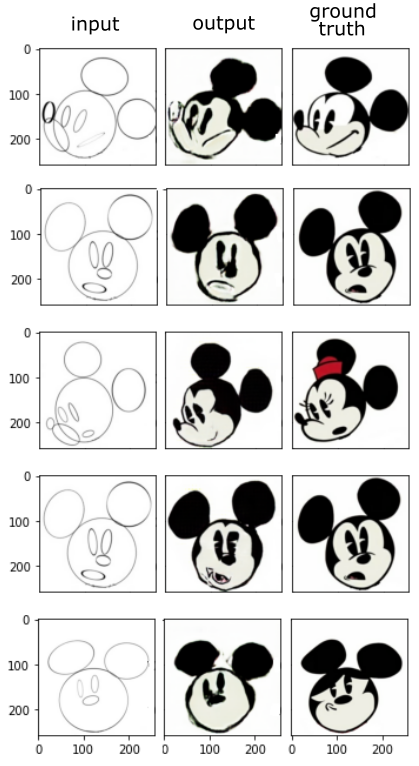}
    \caption{Results of our \textit{shapes2toon} for training on original dataset (without augmentation). Here, \textit{left column} shows input geometric layout, \textit{middle column} shows the generated output based on input. Ground truths are shown in \textit{right column}.}
    \label{fig:orig_res}
\end{figure}

\section{Implementation of \textit{pix2pix} and Experiments}
\label{sec:exp}
To learn a mapping from geometric shapes to actual Mickey Mouse, we used the pix2pix model to train our dataset. The concept of pix2pix is already described in Section~\ref{sec:related}. Fig.~\ref{fig:pix2pix_arch} shows the architecture of our pix2pix implementation. Here, the generator is a \textit{U-Net} based encoder-decoder model~\cite{ronneberger2015unet}. The model begins with a source picture (for example, our layout of geometric shapes) and produces a target image (e.g. cartoon image). This is accomplished by downsampling or encoding the input picture to a bottleneck layer and then upsampling or decoding the bottleneck representation to the output image's size. The U-Net design entails the addition of skip-connections between the encoding and decoding layers, producing a $U$-shape. 

Unlike the standard GAN model, which classifies pictures using a deep convolutional neural network, the Pix2Pix model employs a PatchGAN discriminator. This is a deep convolutional neural network that was developed to identify individual $N\times N$ patches of an input picture as real or false, rather than the full image. We used a patch size of $70\times70$ which is usually effective across a range of image-to-image translation tasks~\cite{brownlee_2019}.

\begin{figure}[htb]
    \centering
    \includegraphics[width=0.95\linewidth]{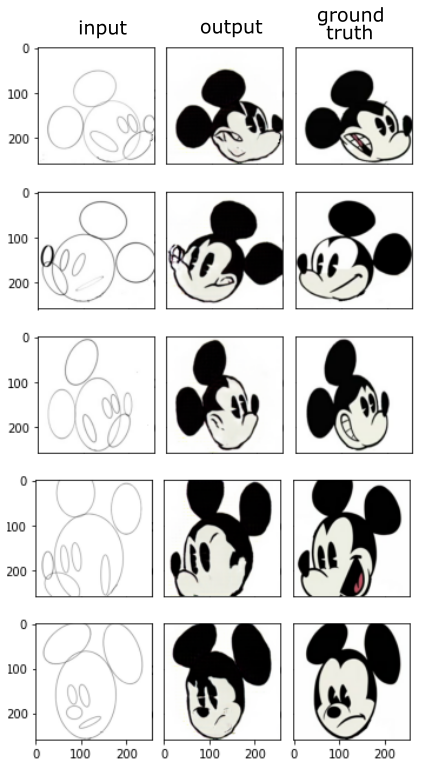}
    \caption{Results of our \textit{shapes2toon} for training on augmented dataset (without augmentation). Here, \textit{left column} shows input geometric layout, \textit{middle column} shows the generated output based on input. Ground truths are shown in \textit{right column}.}
    \label{fig:aug_res}
\end{figure}

\begin{figure}[htb]
    \centering
    \includegraphics[width=\linewidth]{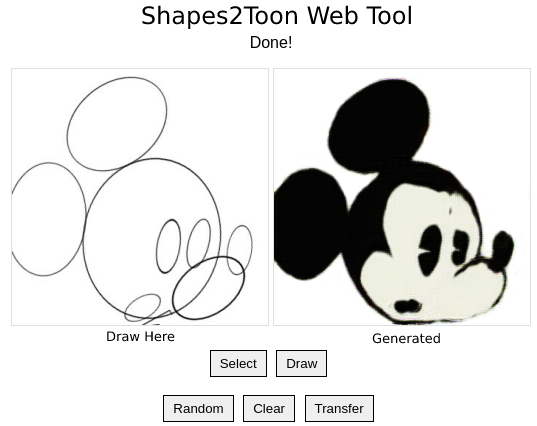}
    \caption{A snap of our \textit{shapes2toon} web application. Here, user can draw shapes'layout via \texttt{draw} button and generate the corresponding Mickey Mouse face via \texttt{transfer}. User can \texttt{clear} the campus any time. For simplicity, a predefined layout can be uploaded using \texttt{select} option. User also can see a sample output from our existing collection of input by clicking \texttt{Random} button.}
    \label{fig:web_tools}
\end{figure}

\subsection{Experimental Setup and Results}

\begin{figure}[htb]
        \centering
        \subfloat[\label{2a}]{%
        \includegraphics[width=75mm,scale=0.1]{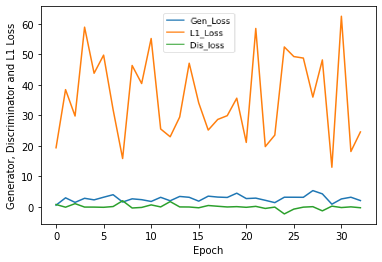}
    }
    \\
        \centering
        \subfloat[\label{2b}]{%
        \includegraphics[width=75mm,scale=0.1]{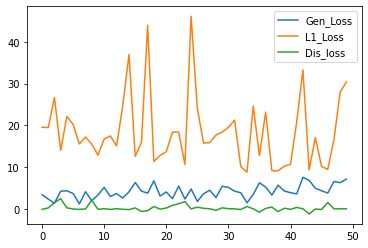}
    }
    \caption{Shows generator, discriminator and $L1$ loss. \textit{For original dataset: (a)} Training process started with a generator loss of $0.395$, discriminator loss of $0.874$ and the $L_1$ loss of $19.831$ and ended with $2.027$, $0.462$, $26.418$ respectively.
    \textit{For our augmented dataset: (b)} Training process started with a generator loss of $3.437$, discriminator loss of $-0.238$, $L_1$ loss of $19.97$ and ended with $7.069$, $-0.124$, $31.46$ respectively. } \label{fig:loss}
\end{figure}

We evaluated our method by conducting experiments on our main dataset as well as the augmented dataset. The main dataset has $500$ images, whereas the augmented dataset contains $7500$ images. The images are scaled to $512\times256$ pixels, with adjoined geometric forms and the original cartoon side-by-side, each of which is $256\times256$ pixels in size. Both datasets were split into $95.8$ percent training and $4.2$ percent testing ratios.

For all datasets, we set the number of filters for the both generator and discriminator, $Ng = 64, Nd = 64$.
We started training the model with a learning rate of $0.0002$ having a batch size of $1$ and, trained the network for $30$ epochs and $50$ epochs for the main dataset and the augmented dataset respectively.
The training and testing process carried out on \texttt{Tesla K80} and \texttt{cuda V10.1} embedded in \texttt{Google Colaboratory}.\\

We perform quantitative and qualitative studies to validate our method. Fig.~\ref{fig:orig_res} and \ref{fig:aug_res} illustrate several qualitative visualizations. For quantitative assessment, we use \textit{Frechet inception distance (FID)} \cite{DBLP:journals/corr/HeuselRUNKH17} in this paper. The reason behind choosing Frechet inception distance, the distribution of generated images is compared to the distribution of a set of real images by FID unlike the earlier inception score which evaluates the distribution of generated images without any ground truth. Lower FID values indicate closer distances between synthetic and real data distributions. The Pix2pix model achieved an FID score of $244.651$ on the main dataset and an FID score of $183.403$ on the augmented dataset. Fig. ~\ref{fig:loss} depicts the training loss graphs on our both datasets.

\begin{figure}[htb]
    \centering
    \subfloat[\label{1a}]{
        \includegraphics[width=\linewidth]{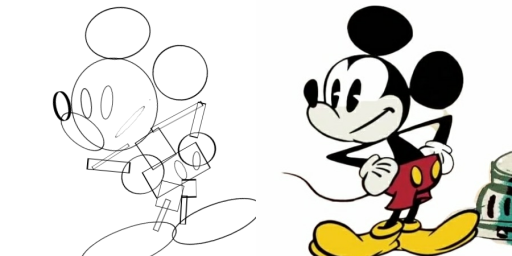}
    }
    
    \centering
    \subfloat[\label{1b}]{%
        \includegraphics[width=0.9\linewidth]{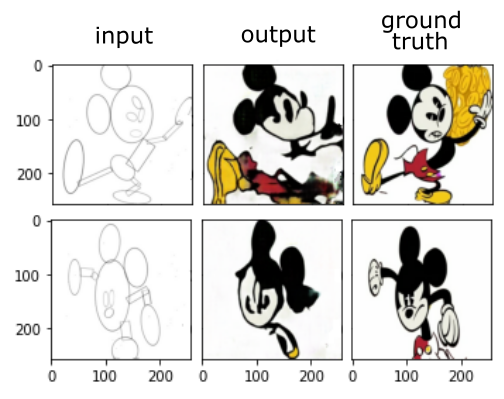}
    }
    \caption{Our attempts to generate full body of Mickey Mouse character. Here (a) shows an example of creating a paired sample for the dataset while considering the full body. Some results of applying pix2pix on full-body dataset are shown in (b).}
    \label{fig:limitation_results}
\end{figure}

\section{Conclusion and Future Works}
\label{sec:con}
In this research, we presented a technique called \textit{shapes2toon} that tries to automate this process by employing a generative adversarial network that mixes geometric primitives (such as circles and ovals) to produce a cartoon character (such as Mickey Mouse) based on the supplied approximation. We were inspired from the fact that drawing basic geometric primitives as an initial approximation of a cartoon is commonly used by people. We generated a collection of geometrically represented cartoon characters from YouTube source to build a paired dataset for this purpose. On our dataset, we used an image-to-image translation approach (\textit{pix2pix}) and described the findings in this paper. The results of the experiments indicated that our system can produce cartoon characters from a geometric form input arrangement. In addition, as a practical application of our research, we presented a web-based utility. I reckon, our technique can be a helpful option for children and novices to learn cartoons.
However, there are some limitations of our work such as dealing with the full body of the cartoon. We already attempted to expand our dataset for including the entire body; however due to the extreme nature of the cartoon bodies and diversity, the results were not satisfactory (see Fig.~\ref{fig:limitation_results}). Moreover, we currently focus on a specific cartoon character and highlight on the facial region only.\\

\textit{Future Works.}
The analysis on the present limitation opens the future avenue to improve our work. Based on the discussed issue we list the following tasks as our future plans.

\begin{itemize}
    \item We plan to expand our dataset by collecting more images of various cartoon character that will be suitable for full body of the cartoon character. We need a robust cartoon figure extraction tool for this purpose.
    \item In future, we want to experiment on different variants of pix2pix and provide necessary ablation studies. We also desire to build a dedicated architecture for our shapes2toon.
    \item Applying different quality assessment is also one of our future targets. We plan to implement \textit{AMT perceptual studies} which are used in many image to image translation methods, such as~\cite{zhang2016colorful}~\cite{quality2004}. 
\end{itemize}

\bibliographystyle{IEEEtran}
\bibliography{ref}


\end{document}